\documentclass[11pt]{article}
\usepackage[final]{acl}

\usepackage{times}
\usepackage{latexsym}
\usepackage[T1]{fontenc}
\usepackage[utf8]{inputenc}
\usepackage{microtype}
\usepackage{inconsolata}
\usepackage{graphicx}
\usepackage{booktabs}
\usepackage{array}
\usepackage{amsmath}
\usepackage{amssymb}
\usepackage{xcolor}
\usepackage{bbm}

\title{AgentDiff: Meaning-Bearing Rewrites Trigger Deeper Divergence\\
than Presentation Changes in LLM Agents}

\author{
Liyun Zhang$^{\heartsuit}$\thanks{Work done during an internship at Tencent.}
\quad
Jiayi Guo$^{\diamondsuit}$ \\
$^{\heartsuit}$School of Information and Software Engineering, UESTC \\
$^{\diamondsuit}$Jacobs School of Engineering, UC San Diego \\
\texttt{jig073@ucsd.edu}
}

\begin{document}
\maketitle

\begin{abstract}
LLM agents should respond to what an input means, not how it is presented. We show that they do not treat these two kinds of variation equally. AgentDiff measures the difference between answer changes caused by meaning-bearing rewrites (paraphrase and synonym substitution) and presentation changes (reordering, formatting, and distractors), while matching perturbation severity. Across 68 model--benchmark--scaffold cells spanning ten LLMs from seven architecture families, three benchmarks, 1{,}530 original questions, and approximately 11{,}150 variants, meaning-bearing rewrites produce a $+19.69$ percentage-point higher inconsistency rate (paired $t=9.58$, $p<0.0001$; 64/68 cells positive). The result is stable under four severity proxies ($+18.9$ to $+20.9$~pp, all $p<0.0001$) and remains $+11.10$~pp on the 48 cells outside the qwen family. A fully held-out qwen2.5-14B-Instruct evaluation then tests the structure of this effect on 1{,}800 new trajectories: the pre-registered capable-and-tractable partition is positive in 3/4 held-out cells and remains sharply separated after pooling (Welch $t=3.81$, $p=9.6\!\times\!10^{-4}$). Trace analysis explains how the difference propagates. Meaning-bearing rewrites preserve the first action but reduce thought similarity from step 2 onward by $5.6$--$10.5$ points and extend the resulting cascade by $0.17$ steps (paired $t=7.69$, $p=2.5\!\times\!10^{-14}$), a pattern we call \emph{stealth divergence}. AgentDiff therefore establishes a reproducible directional robustness gap, identifies the regime in which it is most reliable, and connects final-answer inconsistency to a distinct trajectory-level signature. Code, perturbations, trajectories, and analysis scripts are released for review at \url{https://anonymous.4open.science/r/agentdiff-emnlp-0BB4/}.
\end{abstract}

\section{Introduction}

LLM agents operate downstream of paraphrasers, templates, retrieval systems, and users who express the same request in many forms. Robust behavior therefore requires more than solving a canonical prompt: an agent must preserve its decision when presentation changes while remaining sensitive to content-bearing changes. Yet these two responses are usually pooled into a single robustness score. That aggregation hides the operational distinction that matters most---whether an agent is reacting to meaning-bearing tokens or to the surface form surrounding them.

This distinction is especially consequential for multi-step agents. A small input variation can alter an early thought, tool call, or intermediate state and then propagate through the remainder of the trajectory. Existing evaluations reveal that prompt rewrites and formatting changes can alter model outputs \citep{zhu2024promptbench,sclar2024quantifying}, but they primarily study single-step predictions or aggregate heterogeneous perturbations. Behavioral suites such as CheckList \citep{ribeiro2020beyond} and agent benchmarks such as AgentBench \citep{liu2024agentbench} likewise do not measure a directional gap between meaning-bearing rewrites and presentation changes along complete reasoning trajectories. As a result, we lack both a controlled measurement of this gap and an account of how it develops inside an agent run.

Accuracy alone cannot expose this distinction. Two agents can attain the same benchmark score while reacting very differently to equivalent reformulations: one preserves its answer under layout and wording changes, whereas the other follows a different reasoning path each time the prompt is rendered. Conversely, an answer can remain correct even when the underlying trajectory changes substantially. AgentDiff therefore treats the original run as a paired reference and measures both final-answer inconsistency and step-level displacement. This pairing isolates sensitivity to input form from differences in item difficulty and turns robustness into a property of an agent--task configuration rather than a single aggregate score.

The direction of the gap also carries operational meaning. If presentation changes dominate, normalisation and prompt templating are natural interventions. If meaning-bearing rewrites dominate, the agent is sensitive to how task content is lexicalised even when the requested solution is unchanged. Distinguishing these regimes informs evaluation design, model selection, and the construction of upstream paraphrasing or retrieval components. It also creates a direct bridge between behavioral robustness and trajectory analysis: the final answer measures whether the perturbation matters, while the trace reveals where its effect accumulates.

We introduce AgentDiff to supply both. The study covers 68 cells built from ten LLMs in seven architecture families, three benchmarks---GSM8K \citep{cobbe2021training}, MATH \citep{hendrycks2021measuring}, and HotpotQA \citep{yang2018hotpotqa}---and two agent scaffolds, chain-of-thought \citep{wei2022chain} and ReAct \citep{yao2023react}. Each cell applies the same five operators: paraphrase and synonym substitution target meaning-bearing tokens, while reordering, formatting, and distractor insertion alter presentation. AgentDiff records final answers and full trajectories, matches operator classes by perturbation severity, and audits the result across alternative severity metrics, model families, judges, generators, and nested resampling schemes. We reserve an 11th model, qwen2.5-14B-Instruct, for an independent $n{=}200$-per-cell evaluation and trace-level analysis.

Our contributions are:

\begin{enumerate}
\item \textbf{A severity-controlled directional robustness gap.} Across the 68-cell panel, meaning-bearing rewrites increase inconsistency by $19.69$~pp relative to presentation changes (64/68 cells positive). The result persists across edit distance, token Jaccard, Sentence-BERT cosine, and length-change matching ($+18.9$ to $+20.9$~pp; all $p<0.0001$) and remains $+11.10$~pp on 48 non-qwen cells (\S\ref{sec:severity}).
\item \textbf{A capability-conditional map of where the gap appears.} The pre-registered partition separates capable agents on tractable tasks from weak-model or deep-math settings. A held-out 11th model preserves the predicted direction in 3/4 Group-A cells; the pooled separation is $+10.0$ versus $-1.4$~pp (Welch $t=3.81$, $p=9.6\!\times\!10^{-4}$; \S\ref{sec:held_out}). This identifies a threshold-like operating regime centered on capable agents in tractable tasks.
\item \textbf{A trajectory-level account of stealth divergence.} On 1{,}800 held-out trajectories, meaning-bearing rewrites leave the first action stable but drive a persistent thought-similarity gap from step 2 onward and a $0.17$-step deeper cascade (\S\ref{sec:mechanism}). The combined signal connects final-answer inconsistency to delayed, persistent drift inside the reasoning process.
\item \textbf{A reproducible measurement protocol.} AgentDiff combines matched perturbations, trajectory recording, hierarchical inference, judge replacement, and generator-family calibration in a reusable workflow for evaluating new agents (\S\ref{sec:experiments}--\ref{sec:probe}).
\end{enumerate}

Together, these results turn a broad robustness question into a measurable distinction: meaning-bearing rewrites and presentation changes leave different signatures at both the answer and trajectory levels.

\section{Related Work}
\label{sec:related}

\paragraph{Perturbation robustness for single-step models.} \citet{zhu2024promptbench} introduced PromptBench, a battery of perturbation operators on classification and generation models that found performance drops vary by operator. \citet{ribeiro2020beyond} defined behavioural categories in CheckList that include both meaning-preserving rewrites and meaning-changing edits but operate on the model's own predictions, not on a multi-step trajectory. \citet{gardner2020evaluating} created Contrast Sets --- minimal-pair test items --- that again target single-step predictions. \citet{sclar2024quantifying} showed that prompt formatting alone can shift accuracy by tens of points. AgentDiff extends this line of work by jointly measuring the directional final-answer gap and its step-level propagation across multi-model agents.

These approaches establish two complementary evaluation traditions. Behavioral testing asks whether a model respects invariances specified by the evaluator, while contrast sets measure how local changes move a model's decision boundary. AgentDiff combines these ideas in a paired agent setting: every variant is anchored to the same original question, and the two operator classes are compared after their severity distributions are aligned. The resulting $\Delta$ measures which class more readily moves an agent away from its own reference behavior.

\paragraph{Agent benchmarks and trajectory analysis.} \citet{liu2024agentbench} and \citet{yan2024agentboard} measure end-to-end success on multi-step tasks via AgentBench and AgentBoard, while \citet{yao2025tau} study tool-agent-user interaction in $\tau$-Bench. \citet{yao2023react}, \citet{shinn2023reflexion}, and \citet{madaan2023self} instrument the trajectory itself in ReAct, Reflexion, and Self-Refine. AgentDiff adapts this trajectory-level view to inconsistency and measures cascade depth with both exact-step and TF-IDF cosine alignment (Appendix~\ref{sec:tfidf}).

End-to-end success and perturbation consistency answer different questions. Success records whether an agent eventually reaches the target; consistency records whether equivalent inputs induce the same decision. Trajectory comparison connects them by showing how a local input edit changes the sequence of thoughts and actions before the final answer is produced. This connection is central for CoT and ReAct, where intermediate state can amplify a small input difference across multiple reasoning steps.

\paragraph{Closest prior measurements on multi-step reasoning under perturbation.} \citet{mirzadeh2024gsm} measure GSM8K accuracy under template-level paraphrase and numerical replacement and report drops as large as 65~pp; the present paper measures \emph{directional} inconsistency between meaning-bearing and presentation operators at the trajectory level rather than \emph{undirected} accuracy drop, and uses the cascade-depth statistic introduced in \S\ref{sec:cascade} to split the trajectory's contribution from the final-answer mismatch. \citet{lanham2023measuring} and \citet{turpin2023language} measure faithfulness of chain-of-thought reasoning by intervening on early steps; we measure the cascade footprint of an input-side perturbation through to the final step.

\section{Method: Measurement Framework}
\label{sec:method}

\subsection{Study design}

AgentDiff evaluates a \emph{cell}, defined by one model, one benchmark, and one agent scaffold. The core panel contains ten models from seven architecture families: qwen2.5-3B/7B, llama-3.2-1B/3B, llama-3.1-8B, Mistral-7B, Gemma2-9B, MiMo-v2.5-Pro, Gemini-2.5-Flash, and DeepSeek-V3. This collection spans compact open-weight checkpoints, larger local instruction-tuned models, and commercial API systems. Each model is evaluated with chain-of-thought and ReAct on GSM8K, MATH, and HotpotQA, producing 60 core model--benchmark--scaffold configurations. Eight paired generator configurations extend the analysis to 68 cells.

The three benchmarks provide complementary reasoning environments. GSM8K emphasizes short natural-language arithmetic, MATH requires longer symbolic and competition-style derivations, and HotpotQA combines multi-document evidence across a multi-hop question. CoT exposes a sequence of intermediate thoughts before the final answer. ReAct adds explicit action and observation fields, allowing the same input perturbation to influence both internal reasoning and interaction decisions. Within a cell, the original and every accepted variant use the same model, scaffold, prompt format, and tool interface. Cell sizes range from 20 to 50 original questions, yielding 1{,}530 original runs and approximately 11{,}150 accepted variant runs across the panel.

This crossed design serves two purposes. Model diversity reveals whether the phenomenon recurs across capability levels and architectures; benchmark and scaffold diversity reveals whether it survives changes in reasoning structure. The paired construction then isolates the effect of the perturbation itself: every comparison shares the same underlying task, gold answer, agent configuration, and original reference trajectory.

\subsection{Operator taxonomy}

We label perturbation operators by what they target rather than by whether they change meaning, since paraphrase and synonym substitution are formally meaning-preserving rewrites yet they target meaning-bearing tokens.

\begin{table}[t]
\centering
\footnotesize
\setlength{\tabcolsep}{4pt}
\begin{tabular}{@{}llp{3.0cm}@{}}
\toprule
\textbf{Side} & \textbf{Operator} & \textbf{Targets} \\
\midrule
Meaning-bearing & Paraphrase & Whole-question rewrite \\
Meaning-bearing & Synonym    & Open-class word substitution \\
Presentation    & Reorder    & Token / clause permutation \\
Presentation    & Format     & Whitespace, casing \\
Presentation    & Distractor & Insertion of irrelevant context \\
\bottomrule
\end{tabular}
\caption{Five-operator taxonomy used throughout the paper. All five operators are meaning-preserving rewrites in the formal sense; the labels refer to which tokens the operator targets.}
\label{tab:taxonomy}
\end{table}

All five operators preserve the gold answer; an equivalence judge filters out variants that change the underlying question. The labels distinguish where an operator acts: meaning-bearing rewrites modify open-class content or reformulate the full question, whereas presentation changes alter order, layout, or irrelevant context.

\subsection{Variant construction and validation}

Each original question produces one candidate from every operator in Table~\ref{tab:taxonomy}. Paraphrase and synonym variants are generated with a fixed qwen2.5-3B model. The paraphrase instruction preserves proper nouns, numbers, units, and named entities while changing syntax and non-essential wording. The synonym instruction replaces two or three common verbs or adjectives while preserving domain terms and answer-bearing entities. Generated text is filtered for empty responses, prompt echoes, degenerate length, and unchanged copies; generation is retried up to two times before a deterministic lexical fallback is used.

The three presentation operators are deterministic functions of the original question. Reorder permutes declarative sentences or clauses while anchoring the interrogative sentence. Format converts statements into a bullet or line-oriented representation and places the question on a dedicated line. Distractor inserts an explicitly irrelevant sentence selected deterministically from a fixed pool. For HotpotQA, the same reorder and format logic also applies to the evidence context. A hash of the original question seeds all deterministic choices, so repeated runs reconstruct the same variant.

Validation follows the structure of the operator. A qwen2.5-7B equivalence judge evaluates paraphrase and synonym candidates against the original question and reference answer, accepting a variant when correct solutions yield the same answer. Presentation variants pass a structural invariant check that preserves all numbers, currency and unit symbols, interrogative words, and question marks from the original. Only accepted variants enter the agent evaluation. This pipeline keeps the task identity fixed while allowing the two operator classes to occupy overlapping levels of lexical and embedding-space displacement.

\subsection{The inconsistency rate gap $\Delta$}

For each cell $c$ (a model $\times$ benchmark $\times$ scaffold combination) and each operator $o$, the inconsistency rate is
\begin{equation}
\mathrm{IR}_{c,o} = \frac{1}{N_c}\sum_{i=1}^{N_c}\mathbf{1}\!\left[a_{c,o,i}\ne a^{\mathrm{orig}}_{c,i}\right],
\end{equation}

where $a_{c,o,i}$ is the agent's final answer on the perturbed variant of original question $i$ and $a^{\mathrm{orig}}_{c,i}$ is the answer on the original question. The gap is then
\begin{equation}
\Delta_c = \overline{\mathrm{IR}}_{c,\mathrm{sem}} - \overline{\mathrm{IR}}_{c,\mathrm{sur}},
\end{equation}
where $\overline{\mathrm{IR}}_{c,\mathrm{sem}}$ averages over paraphrase and synonym, and $\overline{\mathrm{IR}}_{c,\mathrm{sur}}$ averages over reorder, format, and distractor.

For every accepted variant, the agent runs from a fresh state and produces a final answer under the same scaffold used for the original. Mathematical answers are compared after normalising the extracted final expression; open-domain answers use benchmark-appropriate equivalence judging. The indicator in Eq.~(1) therefore records a change relative to the agent's own original answer, independent of whether that reference answer matches the benchmark gold label. Accuracy remains available as a separate cell-level capability measure, while IR captures behavioral stability.

A positive $\Delta_c$ means that meaning-bearing rewrites move the agent away from its reference answer more often than presentation changes of matched severity. A negative value indicates the opposite ordering, and a value near zero indicates comparable sensitivity. Averaging operators within each class prevents any single rewrite template from defining the class-level comparison. The principal analysis treats cells as the unit of replication, preserving the model--task--scaffold structure of the experiment.

\subsection{Cascade depth}
\label{sec:cascade}

For each inconsistent variant we compare the original and perturbed agent traces step by step. Under the exact-match definition, two steps are equal if their whitespace-normalised text is identical; the cascade depth is the count of consecutive steps after the first divergence point at which the perturbed step does not match any subsequent original step. We complement this lexical definition with TF-IDF cosine alignment at thresholds 0.3, 0.5, and 0.7 (\S\ref{sec:tfidf}), allowing the analysis to capture semantically similar steps even when their wording differs.

Each stored trace contains the intermediate thought, action, observation, final-answer marker, and total step count. We align original and perturbed traces by step index and record the first divergence, the number of affected downstream steps, and thought similarity at each aligned position. Cascade depth captures persistence: a one-step detour that immediately returns to the reference path receives a shallow value, whereas a perturbation that reshapes the remainder of the run receives a deeper value. Per-step thought similarity complements this count with a continuous view of how far the perturbed trajectory moves from the original.

\subsection{Inferential model}

Our generalisation test uses the pre-registered categorical partition evaluated on the held-out 11th model (\S\ref{sec:held_out}). The partition is defined before the held-out trajectories are analysed and is tested without retuning its threshold, task groups, or judge prompt.

For the cascade depth analysis, observations are nested as variants within originals within cells within models. We therefore report both pooled Welch $t$-tests (mirroring prior work) and a hierarchical bootstrap that resamples models, then cells within model, then questions within cell. Cell-level paired $t$-tests with $K{=}20$ cells per benchmark serve as a sanity check.

Severity-matched analyses use within-cell quantile bins so that meaning-bearing and presentation variants contribute comparable displacement distributions. We report both paired $t$-tests and Wilcoxon signed-rank tests across cells. The held-out partition is evaluated by its mean gap, sign consistency, Welch separation, Mann--Whitney statistic, and Fisher exact test. Together, these tests connect effect magnitude, direction, and replication across independent cells.

\section{Experiments: Evidence for the Directional Gap}
\label{sec:experiments}

\begin{figure}[t]
\centering
\includegraphics[width=\columnwidth]{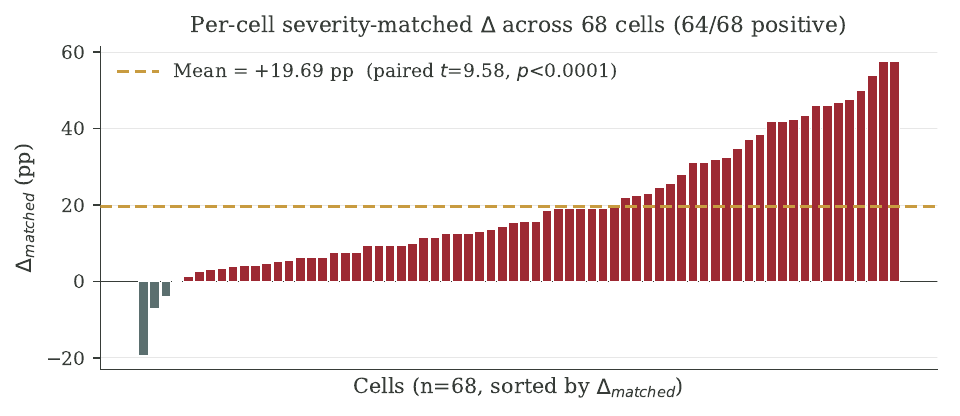}
\caption{Per-cell severity-matched $\Delta$ across 68 cells of the 10-model panel; 64/68 are positive, with a mean of $+19.69$~pp (paired $t=9.58$, $p<0.0001$).}
\label{fig:fig1}
\end{figure}

The experiments establish three linked results. First, the directional gap remains large after matching perturbation severity and changing the severity definition (\S\ref{sec:severity}). Second, a pre-registered capability--tractability partition predicts where the gap appears on a held-out model (\S\ref{sec:held_out}). Third, held-out trajectories reveal stealth divergence: meaning-bearing rewrites preserve the first action but produce deeper and more persistent intermediate drift (\S\ref{sec:mechanism}).

Figure~\ref{fig:fig1} shows that the first result is a panel-wide regularity rather than an average produced by a small set of cells. Sixty-four of 68 matched cell estimates are positive, and the mean gap is $+19.69$~pp. Removing all qwen test models leaves 48 cells spanning Llama, MiMo, Mistral, Gemma2, Gemini, and DeepSeek; their mean gap remains $+11.10$~pp ($p<0.0001$). The shared direction across these families establishes meaning-bearing sensitivity as a recurring property of agent evaluation.

\subsection{The directional gap persists after severity matching}
\label{sec:severity}

Perturbation strength differs across operators, so the primary estimate compares meaning-bearing and presentation variants at matched severity. Normalised Levenshtein distances across approximately 11{,}150 (original, variant) pairs average 0.480 for paraphrase, 0.257 for synonym substitution, 0.284 for reordering, 0.078 for formatting, and 0.485 for distractor insertion. This range supplies overlapping severity levels for a within-cell matched comparison.

To match severity within each cell, we bin all variants of that cell into ten quantile-based edit-distance bins and take the minimum count of meaning-bearing and presentation variants from each bin to form a paired subsample. We then recompute $\Delta$ on the matched subsample. Table~\ref{tab:severity_matched} reports the per-cell distribution.

\begin{table}[t]
\centering
\footnotesize
\setlength{\tabcolsep}{4pt}
\begin{tabular}{@{}lcc@{}}
\toprule
\textbf{Statistic} & $\Delta_{\text{raw}}$ & $\Delta_{\text{matched}}$ \\
\midrule
Mean across 68 cells (pp) & $+18.90$ & $+19.69$ \\
Cells with $\Delta>0$     & 63 / 68 & 64 / 68 \\
Median (pp)               & $+18.75$ & $+15.50$ \\
Paired $t$ vs.\ zero, $t$ & $11.43$ & $9.58$ \\
Paired $t$, $p$           & $<10^{-4}$ & $<10^{-4}$ \\
Wilcoxon, $p$             & $<10^{-4}$ & $<10^{-4}$ \\
\bottomrule
\end{tabular}
\caption{Severity-matched and unmatched inconsistency rate gaps $\Delta$ across 68 cells. Equalising edit-distance distributions within each cell yields a matched gap of $+19.69$~pp, slightly larger than the raw $+18.90$~pp estimate.}
\label{tab:severity_matched}
\end{table}

\begin{figure}[t]
\centering
\includegraphics[width=\columnwidth]{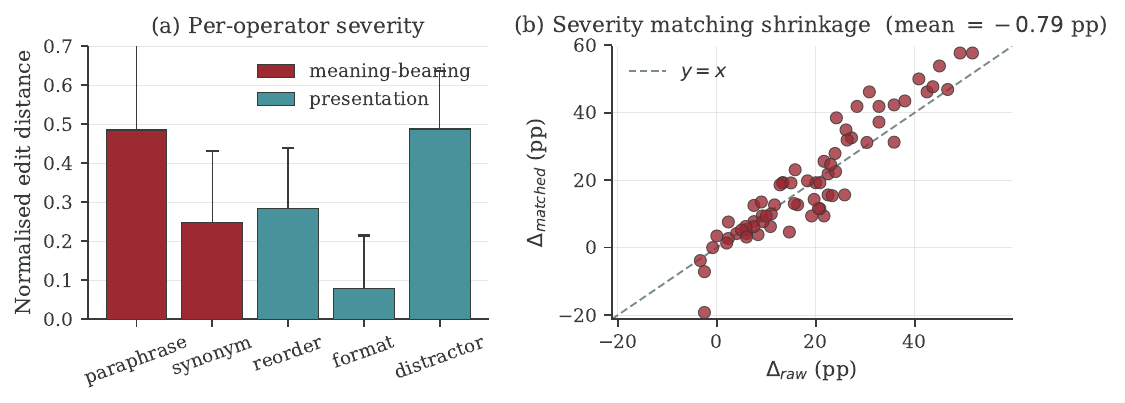}
\caption{(a) Per-operator edit-distance severity (blue: meaning-bearing; orange: presentation). (b) $\Delta_{\text{raw}}$ vs.\ $\Delta_{\text{matched}}$; the dashed line is $y{=}x$. Matching increases the mean gap by $0.79$~pp.}
\label{fig:fig2}
\end{figure}

Severity matching strengthens the directional estimate: mean $\Delta$ increases from $+18.90$ to $+19.69$~pp, and the number of positive cells increases from 63 to 64. The effect therefore reflects operator class beyond raw edit magnitude.

The same comparison is stable across four complementary notions of severity: normalised Levenshtein distance, token-level Jaccard distance, Sentence-BERT cosine distance from the 768-dimensional nomic-embed-text encoder, and absolute prompt-length change. These proxies capture character edits, token overlap, embedding displacement, and prompt expansion, respectively.

\begin{table}[t]
\centering
\footnotesize
\setlength{\tabcolsep}{3pt}
\begin{tabular}{@{}p{3.0cm}cccc@{}}
\toprule
\textbf{Severity proxy} & \textbf{Mean $\Delta$} & $t$ & $p$ & $\Delta>0$ \\
\midrule
Edit distance, normalised      & $+18.94$ & $10.86$ & $<10^{-4}$ & 63/68 \\
Token Jaccard distance         & $+20.91$ & $10.52$ & $<10^{-4}$ & 61/68 \\
Sentence-BERT cosine distance  & $+19.04$ & $9.49$ & $<10^{-4}$ & 64/68 \\
Absolute length-change ratio   & $+20.09$ & $10.35$ & $<10^{-4}$ & 62/68 \\
\bottomrule
\end{tabular}
\caption{Severity-matched $\Delta$ under four complementary severity proxies. Every definition yields a $+18.94$ to $+20.91$~pp gap with $p<10^{-4}$, showing that the directional result is stable across lexical, token, embedding, and length-based matching.}
\label{tab:severity_proxies}
\end{table}

The four matched estimates occupy a narrow two-point range despite measuring substantially different properties of the rewrite. Character distance rewards literal overlap, Jaccard distance tracks token replacement, embedding distance captures global semantic displacement, and length ratio tracks expansion or compression. Their agreement makes the class ordering easy to interpret: at comparable perturbation strength, edits to meaning-bearing wording are consistently more likely to redirect the agent's answer than changes to presentation.

\subsection{Held-out validation on an 11th model: pre-registered capability$\times$tractability partition test}
\label{sec:held_out}

We evaluate the capability-conditional structure on an 11th model that is absent from the main panel and untouched during analysis design: qwen2.5-14B-Instruct. The held-out evaluation runs $n{=}200$ examples across 9 cells (3 benchmarks $\times$ 3 scaffolds, including a \emph{direct} scaffold reserved for exploration), producing 1{,}800 fresh trajectories. We freeze every analysis choice, including thresholds and judge prompts, before reading these results. This design asks a sharper question than another pooled correlation: does the pre-specified capable-and-tractable partition predict the direction of $\Delta$ on a new model?

We apply a $2\times 2$ partition that was pre-registered on the original panel before the held-out data was inspected:
\begin{itemize}
\item \textbf{Group A} = (tier $\in$ \{strong, frontier\}) $\wedge$ (task $\in$ \{shallow\_arith, multi\_hop\}). Qwen2.5-14B is assigned tier=strong by panel-wide accuracy (mean acc 0.81 across its 9 cells); GSM8K is shallow\_arith, MATH is deep\_math, HotpotQA is multi\_hop.
\item \textbf{Group B} = (task $=$ deep\_math) $\vee$ (tier $=$ weak).
\item Mid-tier $\times$ \{shallow\_arith, multi\_hop\}: excluded by design.
\end{itemize}

\begin{table}[t]
\centering
\footnotesize
\setlength{\tabcolsep}{4pt}
\begin{tabular}{@{}llcccc@{}}
\toprule
\textbf{Subset} & \textbf{Grp} & $n$ & \textbf{$\Delta$ (pp)} & \textbf{pos.} & \textbf{Welch $p$} \\
\midrule
Orig. panel (28)         & A & 12 & $+13.0$ & 10/12 & $6\!\times\!10^{-4}$ \\
Orig. panel (28)         & B & 16 & $-1.6$  & 4/16  & --- \\
\midrule
Held-out 14B (6)         & A & 4  & $+1.08$ & 3/4   & --- \\
Held-out 14B (6)         & B & 2  & $+0.38$ & 1/2   & --- \\
\midrule
\textbf{Pooled (34)}     & \textbf{A} & \textbf{16} & $\mathbf{+10.0}$ & \textbf{13/16} & $\mathbf{9.6\!\times\!10^{-4}}$ \\
\textbf{Pooled (34)}     & \textbf{B} & \textbf{18} & $\mathbf{-1.4}$  & \textbf{5/18}  & --- \\
\bottomrule
\end{tabular}
\caption{Pre-registered partition test on the original 28 cot/react cells, 6 held-out qwen2.5-14B cot/react cells, and the pooled set. The held-out Group-A direction is positive in 3/4 cells. Pooling preserves a $+10.0$ versus $-1.4$~pp separation (Welch $t{=}3.81$, $p{=}9.6\!\times\!10^{-4}$; Mann--Whitney $U{=}237.0$, $p{=}1.4\!\times\!10^{-3}$). A pooled Fisher exact test on capability $\times$ sign gives $p{=}4.5\!\times\!10^{-3}$.}
\label{tab:partition}
\end{table}

The held-out design tests a structural prediction, not a new panel average. Group A combines an agent capable of solving the task with benchmarks whose required reasoning remains tractable for that agent. In this regime, meaning-bearing rewrites expose alternative interpretations or intermediate choices while presentation edits are more readily absorbed. The pooled ten-point separation shows that capability and task demands jointly locate the cells in which the directional gap becomes a dependable evaluation signal.

\paragraph{A capability threshold.} Across the full 34-cell pool, accuracy and $\Delta$ correlate at $r{=}+0.32$ ($p{=}0.034$), and capable cells are substantially more likely to have $\Delta>0$ (Fisher $p{=}4.5\!\times\!10^{-3}$). The held-out qwen2.5-14B cells preserve the predicted Group-A direction in 3/4 cases, extending the capability-conditioned pattern to a model excluded from analysis design.

\subsection{Trace-level mechanism: stealth divergence}
\label{sec:mechanism}

The partition test in \S\ref{sec:held_out} establishes when the gap appears. We next use 1{,}800 held-out qwen2.5-14B trajectories to measure how far a perturbation propagates after the agent begins reasoning. Two complementary signals capture this propagation: cascade depth, the number of subsequent steps affected after divergence, and per-step thought similarity, the alignment between original and perturbed intermediate states.

\begin{table}[t]
\centering
\scriptsize
\setlength{\tabcolsep}{4pt}
\renewcommand{\arraystretch}{1.1}
\begin{tabular}{@{}>{\raggedright\arraybackslash}p{0.16\columnwidth}>{\raggedright\arraybackslash}p{0.50\columnwidth}>{\raggedright\arraybackslash}p{0.24\columnwidth}@{}}
\toprule
\textbf{Signal} & \textbf{Observed (sem$-$sur)} & \textbf{Interpretation} \\
\midrule
Cascade depth   & $+0.167$ steps; $t{=}7.69$, $p{=}2.5{\times}10^{-14}$ & deeper propagation \\
Similarity, step 2 & $-0.056$; $t{=}-7.12$, $p{=}2.0{\times}10^{-12}$ & persistent drift \\
Similarity, step 3 & $-0.093$; $t{=}-9.12$, $p{=}1.1{\times}10^{-18}$ & persistent drift \\
Similarity, step 4 & $-0.088$; $t{=}-8.50$, $p{=}1.5{\times}10^{-16}$ & persistent drift \\
\bottomrule
\end{tabular}
\caption{Trajectory-level signature on 1{,}800 held-out qwen2.5-14B runs. Meaning-bearing rewrites produce both a deeper cascade and a persistent reduction in thought similarity.}
\label{tab:mechanism}
\end{table}

\begin{figure}[t]
\centering
\includegraphics[width=0.85\columnwidth]{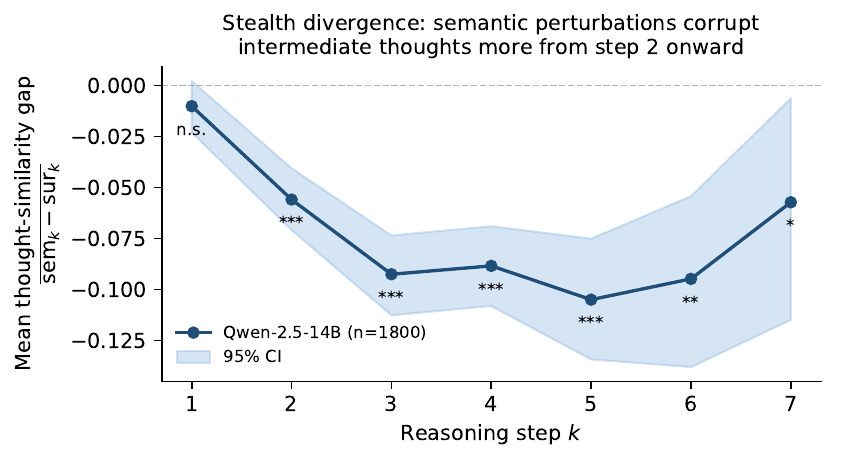}
\caption{Per-step thought-similarity gap (sem $-$ sur) on the qwen2.5-14B trace probe ($n{=}1{,}800$). The two classes share a comparable first step; from step 2 onward, meaning-bearing rewrites produce a persistent similarity gap and a $0.17$-step deeper cascade. This delayed but sustained trajectory shift defines \textbf{stealth divergence}.}
\label{fig:fig6}
\end{figure}

\paragraph{Stealth divergence.} Meaning-bearing rewrites preserve the agent's initial observable action, then move its intermediate thoughts progressively farther from the original trajectory. The similarity gap opens at step 2, persists through later steps, and accompanies a $0.167$-step deeper cascade. This combination of stable onset and sustained internal drift is the signature of stealth divergence.

Stealth divergence explains why final-action monitoring can miss consequential sensitivity. Two runs can begin with the same action and appear behaviorally aligned while their intermediate representations are already separating. Once this displacement persists across several steps, later reasoning operates on a different internal context and the probability of a different answer rises. AgentDiff makes this delayed propagation visible by pairing the answer-level gap with the trajectory-level similarity profile.

\section{Practical Implications}

AgentDiff turns robustness evaluation into a calibrated comparison that can be run before deploying an agent on a target task. A practitioner selects representative questions, generates the five matched variants, and reads $\Delta$ together with the trajectory profile. A large positive gap prioritises evaluation of paraphrase handling and meaning-bearing lexical variation; a shallow trajectory gap points toward local sensitivity, while stealth divergence identifies perturbations whose influence accumulates after an apparently stable first action.

The capability threshold also guides where this measurement is most informative. Once an agent reliably solves the underlying task, consistency under presentation changes becomes a meaningful baseline and the remaining meaning-bearing gap exposes a distinct robustness property. Reporting accuracy, $\Delta$, and trajectory persistence together therefore separates task competence, input invariance, and propagation dynamics---three properties that a single benchmark score conflates.

\section{AgentDiff-Probe v2: Reproducible Measurement Package}
\label{sec:probe}

We package the complete workflow as AgentDiff-Probe v2. Given a calibration set ($\geq 30$ originals $\times$ 5 perturbations $\times$ 2 scaffolds), it generates matched variants, records answer and trace changes, and reports per-cell $\Delta$ with the same severity, judge, and generator-calibration analyses used in this study. The package turns AgentDiff from a fixed benchmark result into a reusable measurement protocol for characterising the robustness profile of a target agent.

\section{Conclusion}
\label{sec:conclusion}

AgentDiff reveals a consistent asymmetry in LLM-agent robustness: after matching perturbation severity, meaning-bearing rewrites change final answers $19.69$ percentage points more often than presentation changes across 68 cells, with 64/68 cells in the same direction. A held-out 11th model confirms the pre-registered capability-conditional structure, separating capable-and-tractable cells from the remaining regime at $p=9.6\!\times\!10^{-4}$ after pooling. The trajectory evidence explains why the asymmetry matters: meaning-bearing rewrites can preserve the first action while inducing deeper, persistent thought drift---stealth divergence. These results establish a concrete target for agent evaluation and a reusable protocol for measuring it.

\section*{Limitations}

This study instantiates AgentDiff with ten panel models, one held-out model, five perturbation operators, and three reasoning benchmarks. As a calibrated measurement protocol, AgentDiff can be extended along each axis: additional model and generator families can map new operating regimes, while human-labelled equivalence and trajectory annotations can refine the perturbation and mechanism measurements. These extensions broaden the empirical map built here while preserving its central comparison between meaning-bearing and presentation variation.

\section*{Ethics Statement}

This work uses three publicly available benchmarks (GSM8K, MATH, HotpotQA) under their respective research-use licences; no human subjects were recruited. Perturbation generation and judging use open-weight or self-hosted LLMs, and the released data contain no personally identifying information. AgentDiff-Probe v2 is intended for research auditing and calibration of agent robustness. Compute usage was approximately 24 CPU-only days on a 48-core, 64~GB RAM workstation plus bounded API allocations for MiMo-v2.5-Pro (${\sim}200$~M tokens), Gemini-2.5-Flash (${\sim}\$2$ USD), and DeepSeek-V3 (${\sim}$\textyen$10$).

\bibliography{references}

\appendix
\section{Supplementary Analyses}
\label{sec:scope_appendix}

\subsection{Model-clustered descriptive regression}

We use the following regression to describe cell-level heterogeneity:
\begin{equation}
\begin{aligned}
\Delta_c ={}& \alpha + \beta_1\,\mathrm{multi\text{-}path}_c \\
&+ \beta_2\,\mathrm{accuracy}_c
+ \beta_3\,\mathrm{ReAct}_c + \varepsilon_c.
\end{aligned}
\end{equation}
Model identity defines the cluster ($K{=}10$). We estimate two-sided $p$-values with 10{,}000 null-imposed Rademacher wild-bootstrap replicates and apply Benjamini--Hochberg correction across the four coefficients.

\begin{table}[h]
\centering
\footnotesize
\setlength{\tabcolsep}{4pt}
\begin{tabular}{@{}lccccc@{}}
\toprule
\textbf{Coefficient} & $\beta$ (pp) & CR1 SE & $t$ & wild $p$ & BH $q$ \\
\midrule
Intercept       & $-2.48$  & $1.79$ & $-1.38$ & $0.108$ & $0.230$ \\
Multi-path      & $+4.31$  & $2.40$ & $+1.80$ & $0.165$ & $0.230$ \\
Accuracy        & $+11.49$ & $5.81$ & $+1.98$ & $0.126$ & $0.230$ \\
ReAct scaffold  & $-1.25$  & $2.74$ & $-0.46$ & $0.626$ & $0.626$ \\
\bottomrule
\end{tabular}
\caption{Cell-level regression with model-clustered wild-bootstrap inference.}
\label{tab:wild_K6}
\end{table}

\subsection{Within-benchmark topology proxies}

We additionally stratify GSM8K by multi-route versus single-route heuristics, MATH by subject-based multi-method versus single-canonical labels, and HotpotQA by multi-evidence comparison versus unique-chain bridge questions. Both GSM8K strata have positive gaps (multi-route $p{=}0.027$; single-route $p{=}0.003$), as does unique-chain HotpotQA ($p{=}0.001$). The between-stratum contrasts are GSM8K $t{=}-1.48$, $p{=}0.155$; MATH $t{=}-0.82$, $p{=}0.426$; and HotpotQA $t{=}+0.05$, $p{=}0.962$.

\begin{figure}[h]
\centering
\includegraphics[width=\columnwidth]{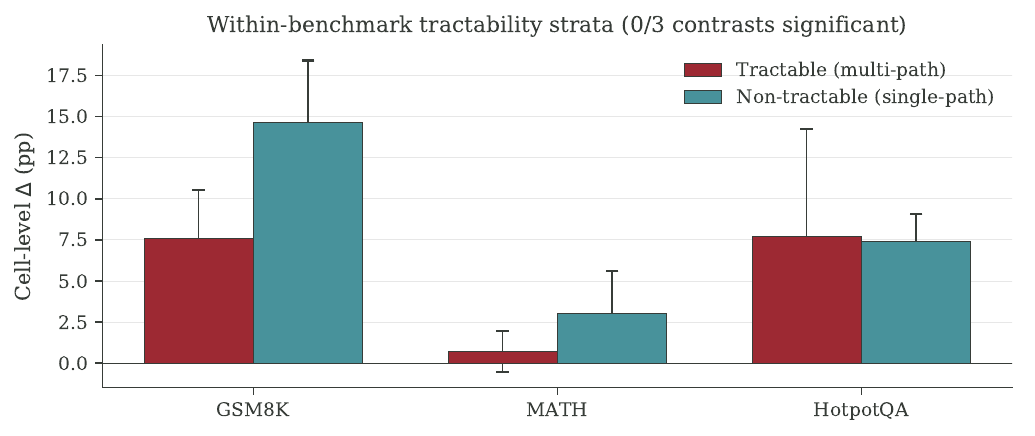}
\caption{Within-benchmark topology strata for GSM8K, MATH, and HotpotQA. Error bars show standard errors across 20 cells per benchmark.}
\label{fig:fig4}
\end{figure}

\subsection{Panel cascade analysis}
\label{sec:tfidf}

We resample model, cell, and question levels for 5{,}000 hierarchical-bootstrap replicates. On inconsistent traces, the cascade-depth gaps are GSM8K $+0.38$ steps (95\% CI $[+0.02,+0.87]$, $p{=}0.035$), MATH $+0.04$ ($p{=}0.90$), and HotpotQA $-0.17$ ($p{=}0.14$). We also align trajectory steps by TF-IDF cosine similarity. The GSM8K gap is $+0.66$ steps at $\cos\geq0.3$ ($p<10^{-3}$), $+0.46$ at $\cos\geq0.5$ ($p=0.003$), and $+0.24$ at $\cos\geq0.7$ ($p=0.139$).

\begin{figure}[h]
\centering
\includegraphics[width=\columnwidth]{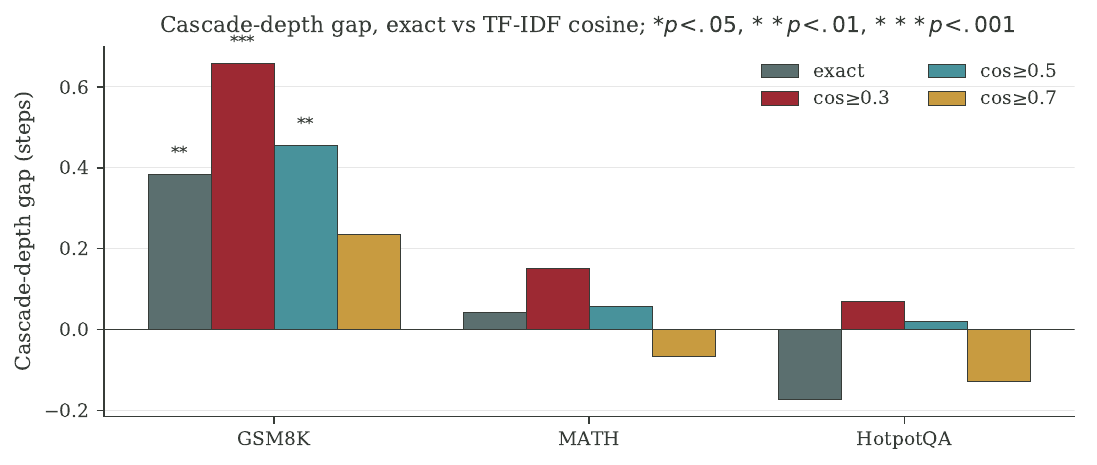}
\caption{GSM8K cascade-depth gap under exact match and TF-IDF cosine alignment.}
\label{fig:fig3}
\end{figure}

\subsection{Cross-family judge audit}
\label{sec:judge}

A MiMo-v2.5-Pro judge re-evaluates a stratified subsample of 1{,}486 decisions originally judged by qwen2.5-7B. Cohen's $\kappa$ is 0.50 overall; 0.50/0.44/0.55 across GSM8K/MATH/HotpotQA; 0.50/0.51 across meaning-bearing/presentation classes; and 0.49/0.51/0.51/0.48/0.52 across the five operators. The second-judge $\Delta$ is positive in 20 of 36 cells on this subsample.

\subsection{Perturbation-generator comparison}
\label{sec:genswap}

We compare cell-level $\Delta$ across qwen2.5:3b, MiMo-v2.5-Pro, and qwen2.5:14b generators on eight paired cells. Qwen2.5:3b vs.\ qwen2.5:14b yields Pearson $r{=}0.794$ ($p{=}0.019$) and Spearman $\rho{=}0.714$ ($p{=}0.047$). Qwen2.5:3b vs.\ MiMo yields $r{=}0.342$ ($p{=}0.41$), $\rho{=}0.143$ ($p{=}0.74$); MiMo vs.\ qwen2.5:14b yields $r{=}0.649$ ($p{=}0.082$), $\rho{=}0.524$ ($p{=}0.18$).

\begin{figure}[h]
\centering
\includegraphics[width=\columnwidth]{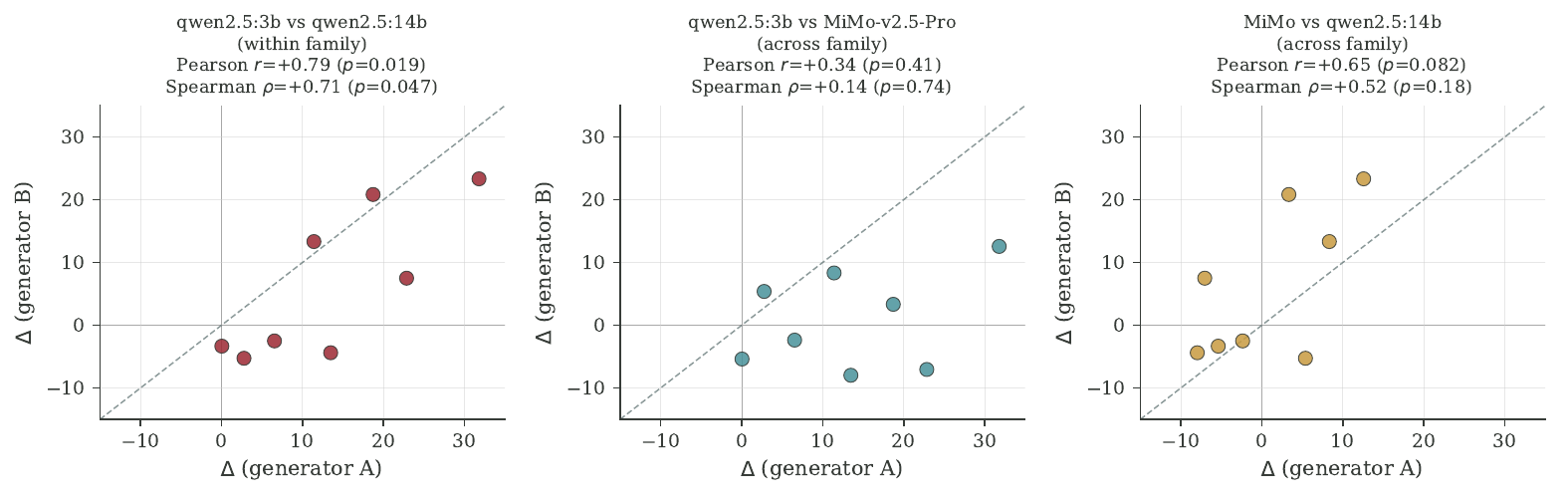}
\caption{Cell-level $\Delta$ across three perturbation generators on eight paired cells.}
\label{fig:fig5}
\end{figure}

\subsection{Additional held-out trace probes}

Two additional pre-registered probes measure divergence onset and self-correction. Meaning-bearing rewrites begin diverging $0.156$ steps later ($t{=}7.40$, $p{=}2.1\times10^{-13}$). Self-correction rates are 2.40\% for meaning-bearing and 2.00\% for presentation variants (Fisher $p{=}0.230$).

\end{document}